# t-PINE: Tensor-based Predictable and Interpretable Node Embeddings

Saba A. Al-Sayouri
Systems Science and Industrial Engineering
State University of New York at Binghamton
Email: ssyouri1@binghamton.edu

Ekta Gujral
Computer Science and Engineering
University of California Riverside
Email: egujr001@ucr.edu

Danai Koutra
Computer Science and Engineering
University of Michigan, Ann Arbor
Email: dkoutra@umich.edu

Evangelos E. Papalexakis
Computer Science and Engineering
University of California Riverside
Email: epapalex@cs.ucr.edu

Sarah S. Lam
Systems Science and Industrial Engineering
State University of New York at Binghamton
Email: sarahlam@binghamton.edu

*Abstract*—Graph representations have increasingly grown in popularity during the last years. Existing representation learning approaches explicitly encode network structure. Despite their good performance in downstream processes (e.g., node classification, link prediction), there is still room for improvement in different aspects, like efficacy, visualization, and interpretability. In this paper, we propose, t-PINE, a method that addresses these limitations. Contrary to baseline methods, which generally learn explicit graph representations by solely using an adjacency matrix, t-PINE avails a multi-view information graph—the adjacency matrix represents the first view, and a nearest neighbor adjacency, computed over the node features, is the second view— in order to learn explicit and implicit node representations, using the Canonical Polyadic (a.k.a. CP) decomposition. We argue that the implicit and the explicit mapping from a higher-dimensional to a lower-dimensional vector space is the key to learn more useful, highly predictable, and gracefully interpretable representations. Having good interpretable representations provides a good guidance to understand how each view contributes to the representation learning process. In addition, it helps us to exclude unrelated dimensions. Extensive experiments show that t-PINE drastically outperforms baseline methods by up to $158.6\%$ with respect to Micro-F1, in several multi-label classification problems, while it has high visualization and interpretability utility.

## I. INTRODUCTION

Graphs are widely used to encode relationships in real-world networks, such as social networks, co-authorship networks, biological networks, to name a few. Among the various approaches that have been proposed for network analysis, representation learning has gained significant popularity recently. Representation learning techniques [13], [8], [18] primarily aim to explicitly learn a unified set of representations in a completely unsupervised or semi-supervised manner, which ultimately can generalize across various tasks, such as link prediction [10], node classification [1], and recommendation [22]. However, since the "one-size fits all" approach is adopted in the representation learning context, the explicit learning does not guarantee that the representations would convey enough information crucial for different downstream tasks.

The ideal representations are those that capture both explicit and implicit network connectivity patterns. In real-world networks, the majority of nodes are implicitly related to each other despite missing edges. Existing representation learning approaches [13], [8], [18] can preserve the explicit connections in a graph well, but barely capture the implicit wide-spread connections. Taken together, we argue that since the observed graph does not reflect the actual existing implicit network structure, which ultimately compromises downstream task performance.

Aside from the shallow models, which leverage first-order connectivity, there has been work, where explicit and implicit representations are learned [14]. Although [14] learns representations that can better preserve network structure than prior work [13], [8], [18], it still relies on a family of representation sets, where each set encodes a different scale of relationships (e.g., family, friends, classmates) and those are concatenated to form the final representation. Here we instead seek to learn representations that *jointly* capture connectivity patterns at different scales, and thus, successfully generalize over various downstream processes.

In this paper, we propose, t-PINE, a method (illustrated in Figure 1) that overcomes the limitations of existing representation learning approaches, by using the Canonical Polyadic (CP) or CANDECOMP/PARAFAC decomposition [3] to learn useful, highly predictable, and gracefully interpretable representations. To summarize, our contributions are:

1) **Systematic Exploration of Implicit Higher-order Proximities:** As we argue earlier, the adjacency matrix captures explicit connectivity patterns, which scarcely encode the full network underlying structure. We augment that with another important information view: the $K$-Nearest Neighbors view, which can convey implicit relationships. In order to capture implicit relationships at different scales, we systematically explore higher-order proximities using the intuition of the established $K$-nearest neighbor (a.k.a. $K$-NN) algorithm [15].

2) **Multi-View Representation Learning:** We propose t-PINE, a method that uses CP decomposition to learn rich, useful, highly predictable, and interpretable representa-tions.
3) **Interpretability:** Contrary to the existing representation learning baseline techniques, by using CP decomposition, we can easily interpret the representations under the guidance of factor matrices, and thereby further understand and improve the evaluation results.
4) **Experiments:** We extensively evaluate t-PINE on multi-class classification problems using real-world datasets from different domains. We also benchmark against three existing representation learning approaches.

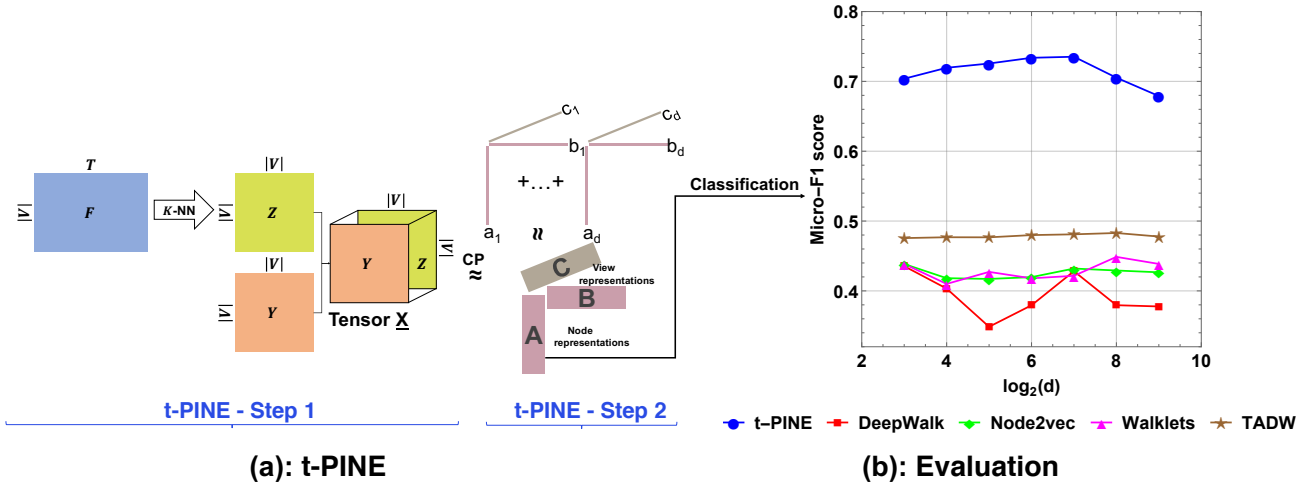

**Figure. 1:** The proposed method and its evaluation. **(a):** t-PINE method. Step 1: Indicates the formation of tensor $\underline{\mathbf{X}}$ using the two views: the adjacency and $K$-NN matrices. Step 2: Shows representation generation process using CP decomposition. **(b):** Evaluation process for t-PINE and comparison of its performance against baselines on the WebKB dataset (Section VA). t-PINE drastically outperforms baseline methods.

## II. RELATED WORK

Here we briefly discuss the two parts of our related work: (1) Conventional representation learning; and (2) Tensor-based representation learning.

**Conventional Representation Learning.** There has been a recent surge of interest in developing the representation learning techniques. In particular, [13], [8], [18] learn node representations by leveraging the explicit connections adjacency matrices convey. Despite their good predictive performance, they still fail to encode the implicit unobserved connectivity patterns. This consequently compromises their generalizability across various downstream tasks. On the other side of the spectrum, [14] sought to capture both the explicit and the implicit connections by extending the DeepWalk [13] model by learning a series of representation corpora. However, for the best performance, the family of representation sets should be concatenated. This renders [14] to lack the capability of *jointly* capturing relationships at different scales. Another representation learning approach has been presented by [21], where an adjacency and a textual information matrices have been employed to learn representations, via matrix factorization.. Table I presents a qualitative comparison of conventional representation learning related work vs. t-PINE. In this paper, we strive to overcome the limitations of the existing representation learning state-of-the-art methods, by: (1) Achieve a high satisfactory predictive performance; (2) Jointly learn explicit and implicit connectivity patterns; and (3) Generate easily interpretable representations.

**Tensor-based Representation Learning.** To the best of our knowledge, this paper is the first to demonstrate the use of tensor factorization techniques in the context of network representation learning. Tensor decomposition captures the node relations via low-dimensional latent components. An extensive overview of tensor decomposition methods is given by [9]. There are very few research work that have been initiated in the NLP domain to learn word representations using tensor decomposition: [6] proposes an algorithm to generate word embeddings of prepositions using 3-mode tensor decomposition; [19] presents an algorithm to investigate three-way co-occurrences using non-negative tensor factorization. Aside from word representations, Cao *el. at.* [2] propose a constrained tensor factorization algorithm; t-BNE, to learn brain network representations. This algorithm uses side information guidance to find an optimal set of features for brain graph classification. However, these approaches are not completely related to our work for the following reasons: (1) The word representation-related techniques are only suitable for the process of feature creation; and (2) These methods capture only simple connectivity patterns than those that emerge in large-scale real-world networks.

## III. PROBLEM FORMULATION

### A. Preliminary Definitions

A tensor or a multi-view graph is a higher order generalization of a matrix. We call tensor $\underline{\mathbf{X}} \in \mathbb{R}^{I \times J \times L}$, a three mode tensor, where modes are the numbers of indices used to index the tensor. Due to the space constraint, we refer the reader to a well-known study on tensors and their applications [9], [12].

TABLE I: Qualitative comparison of conventional representation learning related work vs. t-PINE.

| Approach | Consistent Performance | Preserve Explicit Connections | Preserve Implicit Connections | Interpretability |
|---|---|---|---|---|
| DeepWalk | ✗ | ✓ | ✗ | ✗ |
| node2vec | ✗ | ✓ | ✗ | ✗ |
| Walklets | ✗ | ✓ | ✓ | ✗ |
| TADW | ✗ | ✓ | ? | ✗ |
| **t-PINE** | ✓ | ✓ | ✓ | ✓ |

In Table II, we list the symbols and definitions used throughout the paper. Since we focus on interpretability, we use one of the most widely used tensor decomposition: Canonical Polyadic (CP) or CANDECOMP/PARAFAC decomposition [3]. Consider an example of 3-mode tensor $\underline{\mathbf{X}} \in \mathbb{R}^{I \times J \times L}$ data of Amazon reviews [17] with modes: *Users*, *Product* and *Word*. The $R$-latent component CP decomposition expresses as the summation of rank-1 tensors, i.e., a summation of outer products of three vectors, as follows:

$$\underline{\mathbf{X}} \approx \sum_{r=1}^{R} \mathbf{A}(:,r) \circ \mathbf{B}(:,r) \circ \mathbf{C}(:,r) \quad (1)$$

where $\mathbf{A} \in \mathbb{R}^{I \times R}, \mathbf{B} \in \mathbb{R}^{J \times R}$ and $\mathbf{C} \in \mathbb{R}^{L \times R}$ correspond to factor matrices of users, product, and word in modes, respectively.

TABLE II: Symbols and definitions

| Symbol | Definition |
|---|---|
| $G = (V, E)$ | Unweighted and undirected graph |
| $V$ | Set of vertices (nodes) |
| $E$ | Set of edges |
| $\underline{\mathbf{X}}, \mathbf{X}, \mathbf{x}$ | Tensor, Matrix, Column vector |
| $\mathbb{R}$ | Set of real numbers |
| $\circ$ | Outer product |
| $\|\mathbf{A}\|$ | Frobenius norm |
| $\mathbf{X}(:,r)$ | $r^{th}$ column of $\mathbf{X}$ |
| $\mathbf{X}(r,:)$ | $r^{th}$ row of $\mathbf{X}$ |
| $\mathbf{x}(r)$ | $r^{th}$ element of $\mathbf{x}$ |
| $L$ | Number of views |
| $R$ | Rank of tensor $\underline{\mathbf{X}}$ |
| $d$ | Representation dimensionality |
| $\mathbf{A}, \mathbf{B}, \mathbf{C}$ | Decomposition factor matrices |
| $K$ | Number of nearest nodes |
| $\mathbf{Y}$ | Adjacency matrix/view |
| $\mathbf{F}$ | Textual information (feature) matrix/view |
| $\mathbf{Z}$ | $K$-NN matrix/view |
| $T$ | Number of columns of matrix $\mathbf{F}$ |
| $\mathbf{D}$ | Dot product matrix |
| $\mathbf{N}$ | Norm matrix |

### B. Problem Definition

**Information Network.** An information network is defined as a graph $G = (V, E)$, where $V$ represents the set of nodes connected together by a set of edges $E$. In this paper, we focus on multi-view undirected, unweighted information graphs that describe the same set of nodes to learn network representations. Throughout the paper, we use the terms "view", "matrix", and "layer" interchangeably. For each network, we use two views; (1) The adjacency matrix; a squared node-by-node matrix, which we refer to as, $\mathbf{Y} \in \mathbb{R}^{V \times V}$; and (2) Feature matrix, which is denoted as, $\mathbf{F} \in \mathbb{R}^{V \times T}$. Consider a document classification dataset (e.g., Citeseer publication), so that, the adjacency matrix ($\mathbf{Y}^{PID \times PID}$) captures the relation between various research documents classified in six fields like AI, HCI, ML, IR, DB, and Agents. $PID$ refers to the document's unique identification number. Here in our case, $|PID| = |V|$. The $\mathbf{Y}$'s corresponding textual feature binary matrix is represented by $\mathbf{F}^{PID \times words}$, where $|words| = |T|$. The values represent the presence or absence of a specific word in that document.

**Learning Large-scale Network Representations.** Given a large-scale, multi-view information network, $G = (V, E)$, the problem of learning a graph $G$'s node representations strives to preserve the network explicit and implicit connections, while mapping each node $v \in V$ from a higher-dimensional feature space to a lower-dimensional feature space $\mathbb{R}^d$ using a mapping function, $f_G: V \to \mathbb{R}^d$, where $d \ll |V|$. For simplicity, we assume the tensor's rank $R$ is given. Generally, when tensor decomposition (TD) is used in a multi-label classification context, the tensor rank is set to $R =$ number of classes. That is, the TD soft assignments are used to assign each node to its corresponding class. Instead in this study, we employ CP decomposition for representation learning, therefore, we set $R = d \ll |V|$, where $d$ is a parameter specifying the number of dimensions of our representation vector. The problem we solve is as follows:

> **Given** a 3-dimensional tensor $\underline{\mathbf{X}} \in \mathbb{R}^{V \times V \times L}$, where the first view represents an adjacency matrix $\mathbf{Y} \in \mathbb{R}^{V \times V}$, and the other view indicates its corresponding $K$-NN matrix $\mathbf{Z} \in \mathbb{R}^{V \times V}$,
> **Jointly Learn** an implicit and explicit representation of each vertex $v \in V$ in tensor $\underline{\mathbf{X}}$ under the hood of CP decomposition.

## IV. PROPOSED METHOD

In order to successfully map the nodes of a multi-view graph $G$ from a high-dimensional to a low-dimensional vector space using a mapping function, $f_G: V \to \mathbb{R}^d$, where $d \ll |V|$, we need to concurrently preserve the observed explicit and the unobserved implicit connections, which all the state-of-the-art methods fail to address. Further, In order to overcome the lack of interpretability, the prime limitation of the existing representation learning methods, we here propose, t-PINE, a mathod that learns rich representations using CP decomposition. Below, we describe the two steps of t-PINE.

**Step 1: Systematic Exploration of Implicit Higher-order Proximities.** In the context of representation learning, the *first-order* [7], and the *second-order* [7] proximities have been the key two proximity measures representation learning methods pursue to preserve using adjacency matrix. The *first-order* [7] is used to preserve the explicit network structure [18], while the *second-order* [7] is employed to preserve the implicit network structure by claiming that the nodes that share similar neighbors should be mapped close to each other in the low-dimensional vector space [18], [8]. However, the predictive performance of [18], [8] indicates that utilizing these two proximity measures is insufficient to reconstruct the original network. Therefore, a serious need to more effectively encode network explicit and implicit relationships has emerged.

To this end, in this paper, we propose to *jointly* learn network representations that can effectively generalize across downstream processes, by leveraging a multi-view information graph, using CP decomposition. The first view is the adjacency matrix, $\mathbf{Y} \in \mathbb{R}^{V \times V}$, while the second view is the feature matrix, $\mathbf{F} \in \mathbb{R}^{V \times T}$. However, as shown in Figure 1, we process the feature matrix $\mathbf{F}$ using the $K$-order implicit proximity exploration algorithm (shown in Algorithm 1), where $K$ represents the number of nearest nodes we specify, using the intuition of the $K$-NN algorithm [15] for two reasons: (1) To systematically explore implicit higher order proximities that capture network connectivity patterns, since the direct use of matrix $\mathbf{F}$ does not convey the implicit network connections (proven by TADW's [21] performance in Section V-C); and (2) To make the size of the second view obeys with the size of the first view. We call the feature matrix, $\mathbf{F} \in \mathbb{R}^{V \times T}$, after being processed by the $K$-NN algorithm [15], the $K$-NN matrix, $\mathbf{Z} \in \mathbb{R}^{V \times V}$. Since *cosine* similarity has been a very efficient metric to capture proximity in sparse matrices (e.g., matrix $\mathbf{F}$) [4], we adopt it, as our distance metric.

---

**Algorithm 1:** $K$-order Implicit Proximity Exploration

**Input:** $\mathbf{F} \in \mathbb{R}^{V \times T}$, $K$
**Output:** $\mathbf{Z} \in \mathbb{R}^{V \times V}$
1: $\mathbf{N} = (\sqrt{\sum_v^V |T_v|^2}) * (\sqrt{\sum_v^V |T_v|^2})^T$;
2: $\mathbf{D} = TT'$
3: $\mathbf{Z} = \mathbf{D}/\mathbf{N}$;
4: $\mathbf{Z} = \mathbf{Z} - diag(\mathbf{Z})$
5: **for** $i \leftarrow 1$ to $K$ **do**
6:   $[-, idx] = max(\mathbf{Z})$;
7:   **for** $j \leftarrow 1$ to $V$ **do**
8:     $\mathbf{Z}(j, idx(j)) = -1$;
9:   **end for**
10: **end for**
11: $\mathbf{Z}(\mathbf{Z} > 0) = 0; \mathbf{Z}(\mathbf{Z} == -1) = 1$;
12: **return** $Z \in \mathbb{R}^{V \times V}$.

---

**Step 2: Multi-View Representation Learning.** After we conduct *Step 1*, we now stack up the two views: $\mathbf{Y} \in \mathbb{R}^{V \times V}$ and $\mathbf{Z} \in \mathbb{R}^{V \times V}$ to form the tensor $\underline{\mathbf{X}}$. We decompose the tensor $\underline{\mathbf{X}}$ to compute the $d$ latent components: node representations, using CP decomposition. Below, we show the mathematical formulation of CP decomposition:

$$\mathcal{L} \approx \min ||\underline{\mathbf{X}} - \mathbf{A} \circ \mathbf{B} \circ \mathbf{C}||_F^2 \quad (2)$$

where the factor matrices $\mathbf{A} \in \mathbb{R}^{V \times d}$ and $\mathbf{B} \in \mathbb{R}^{V \times d}$ shares partially similar information as the $\underline{\mathbf{X}}$ is symmetric in the first layer. The factor matrix $\mathbf{C} \in \mathbb{R}^{L \times d}$ indicates the contribution of each *view* or *layer* of tensor $\underline{\mathbf{X}}$ to the learned latent representations. $d$ is a parameter specifying the number of dimensions of our feature representation.

As mentioned earlier, since we employ CP decomposition for representation learning, we set $R = d \ll |V|$. In order to solve the CP decomposition problem (shown in Eq.(2)), we use the Alternating Least Squares (ALS) algorithm [3]. The key idea behind ALS algorithm is to optimize one factor matrix at a time, while fixing all the other factor matrices. It repeatedly iterates over the factor matrices and updates them, as shown in Eq.(3) - Eq.(5), until the objective function stops changing across successive iterations.

$$\mathbf{A} \leftarrow \min ||\underline{\mathbf{X}} - \mathbf{A}(\mathbf{C} \circ \mathbf{B})^T||_F^2 \quad (3)$$

$$\mathbf{B} \leftarrow \min ||\underline{\mathbf{X}} - \mathbf{B}(\mathbf{C} \circ \mathbf{A})^T||_F^2 \quad (4)$$

$$\mathbf{C} \leftarrow \min ||\underline{\mathbf{X}} - \mathbf{C}(\mathbf{B} \circ \mathbf{A})^T||_F^2 \quad (5)$$

**Complexity Analysis and Convergence.** In t-PINE, the procedure of computing the $\mathbf{Z} \in \mathbb{R}^{V \times V}$ graph, takes $O(|V|^2 + K * |V|)$ time. The complexity of each iteration of minimizing $\mathcal{L}$ is $O(NNZ(\underline{\mathbf{X}})) + O(|V|^2 + K|V|)$, where $NNZ()$ indicates the number of non-zero entries. . The ALS algorithm converges to a good approximation error within $50 - 60$ iterations.

## V. EXPERIMENTS

In this section, we aim to answer the following questions:
(**Q1**) How does t-PINE perform in multi-label classification compared to baseline representation learning approaches?
(**Q2**) Does the incorporation of the $K$-NN matrix of a serious need to boost the multi-label classification performance?
(**Q3**) How does CP decomposition help to learn more useful, highly predictable, and gracefully interpretable representations?
(**Q4**) How robust t-PINE is to parameter settings? Before we answer these questions, we provide an overview of the datasets and the baseline representation learning algorithms that we use in our evaluation.

### A. Datasets

Table III provides a brief description of the real-world datasets that we use in our experiments. The reason why we choose these networks (e.g., language, citation, and terrorism), is the ease of obtaining the corresponding textual information matrices: $K$-NN matrices along with the adjacency matrices.

### B. Baseline Techniques

Our experiments evaluate the node representations obtained through t-PINE on standard, non-trivial multi-label classification problems. Here, we compare t-PINE's performance against the following representation learning methods:

- **DeepWalk [13]:** A graphical representation learning algorithm. It has two steps: (1) Neighborhood generation using small truncated random walks, and (2) Representation learning and generation using the SkipGram model.

TABLE III: A brief description of evaluation datasets. Number of edges in $K$-NN matrix varies by $K$. The acronym TFIDF stands for: term frequency-inverse document frequency.

| Dataset | # Vertices | # Edges in $G_Y$ | $T$ | # Edges in $G_Z$ | # Labels | Network Type | Feature Type |
|---|---|---|---|---|---|---|---|
| Wikipedia [21] | 2,405 | 35,962 | 4973 | 149,053 | 20 | Language | TFIDF info |
| WebKB [11] | 877 | 5,168 | 1703 | 36,466 | 5 | Citation | Unique words |
| CiteSeer [21] | 3,312 | 9,464 | 3703 | 49,680 | 6 | Citation | Unique words |
| Terrorist [23] | 848 | 16,392 | 1224 | 82,048 | 4 | Terrorism | Relations |

- **node2vec [8]:** An algorithm that was inspired by DeepWalk [13]. In order to capture network unique connectivity patterns; homophily and structural equivalence, it employs $2^{\text{nd}}$ order biased random walks. To control the search space, two search parameters were used; return parameter, $p$, which controls the likelihood of immediately revisiting the same node in the walk, and in-out parameter, $q$, which allows the search to change between the inward and the outward nodes.
- **Walklets [14]:** This extension of DeepWalk [13] models multi-scale dependencies among nodes in a graph. It proposes to skip some nodes in the random walks employed for neighborhood generation.
- **TADW [21]:** An algorithm that employs the adjacency graph and its corresponding feature textual information matrix, $\mathbf{F} \in \mathbb{R}^{V \times T}$, to learn network representations.

In contrast to t-PINE that *jointly* learns implicit and explicit connectivity patterns, we choose to compare t-PINE's with **DeepWalk** [13], and **node2vec** [8], where both focus on preserving network explicit relationships. In addition, we contrast t-PINE's performance against **Walklets** [14], which *individually* learns implicit and explicit connections using a series of representation corpora, each of which captures higher order relationship of the adjacency matrix. It achieves the best performance, when a series of representation corpora are concatenated. We also compare t-PINE against **TADW** [21], because it employs the adjacency matrix along with its corresponding feature textual information matrix, $\mathbf{F} \in \mathbb{R}^{V \times T}$, to learn rich network representations.

**Experimental Setup.** For [13], [8], [14], we set the number of walks per node to $10$, the walk length to $80$, the neighborhood size to $10$, and the number of dimensions of the feature representation $d = 128$. For node2vec, we set the return parameter $p = 1$, and the in-out parameter $q = 1$, in order to capture the homophily, and the structural equivalence connectivity patterns, respectively. For Walklets, we set the feature representation scale, $\pi = 2$, which captures the relationships captured at scale $2$. For TADW, we set the regularization parameter, $\lambda$, to $0.2$, the text rank, $TR$, to $200$, and $d = 128$.

For t-PINE parameter settings, we vary $K$ by dataset. That is, we use the $k$ value that achieves the best performance for each dataset. We set $d = 128$, in line with the values used for DeepWalk, node2vec, Walklets, and TADW.

### C. Experimental Results

In this section, we present our experimental results of the multi-label classification task using the datasets described earlier in Section V.

**Q1. Multi-label Classification**

**Setup.** The multi-label classification problem is a single-graph canonical task, where each node in a graph is assigned to a single or multiple labels from a finite set of labels, $L$. In our study, we feed the learned feature representations to a one-vs-rest logistic regression classifier with the default L2 regularization. The multi-label classification is a challenging task, especially when the finite set of labels, $L$, is large, or the fraction of labeled nodes is small [16]. We report the mean Micro-F1 score results of the 10-fold cross validation.

**Results.** Tables IV presents t-PINE's performance and its gain over the state-of-the-art techniques across different multi-label classification problems, using the mean Micro-F1 score. For each dataset, we list the scores that correspond to a specific $K$. We choose the $K$ value that achieves the best scores. We vary $K$ from $2$ to $40$. Below we discuss the performance per dataset.

**Wikipedia.** We observe that t-PINE outperforms the baseline methods across different training percentages of labeled data, except for DeepWalk when the training percentage of labeled data $= 50\%$, with respect to the Micro-F1 score. It achieves at most a gain of $220.2\%$ when the labeled data is sparse ($10\%$). Further, despite the fact that TADW is the most competitive state-of-the-art method, it achieves significantly lower accuracy than t-PINE. We argue that TADW's high predictive accuracy is attributed to the high predictive power of the support vector machine (SVM) classifier TADW employed for training and prediction. Yet, in this paper, we use a simple logistic regression classifier, which is proven not to be as powerful as SVM. The same reasoning applies to the rest datasets.

**WebKB.** t-PINE drastically outperforms the baseline methods across different training percentages of labeled data. Interestingly, using t-PINE allows us to uncover the unique local and global connectivity patterns baselines are incapable of. In particular, when the labeled data is sparse ($10\%$), t-PINE outperforms state-of-the-art methods by at most $74.4\%$ with respect to the Micro-F1 score.

**CiteSeer.** t-PINE outperforms the baseline methods across different training percentages of labeled data. Remarkably, it surpasses the baselines most when the labeled data is sparse ($10\%$) by at most $158.6\%$ with respect to the Micro-F1 score.

**Terrorist.** We observe that t-PINE outperforms the baseline methods across different training percentages of labeled data. Conspicuously, we note that the baselines are almost on par with t-PINE, which can be rooted in the fact that Terrorist network structure is easy to capture and it highly corresponds to the label information.

TABLE IV: Micro-F1 scores for multi-label classification problems on various datasets. The representation feature space has 128 dimensions. Numbers where t-PINE outperforms other baselines are bolded. For each dataset, we report the used $K$ between two parentheses that yields the best performance. Remarkably, tensor-based embeddings better preserve network structure, which ultimately, improves task performance.

| Algorithm | Wikipedia ($K=8$) | | | WebKB ($K=40$) | | | CiteSeer ($K=15$) | | | Terrorist ($K=25$) | | |
|---|---|---|---|---|---|---|---|---|---|---|---|---|
| | 10% | 50% | 9% | 10% | 50% | 90% | 10% | 50% | 90% | 10% | 50% | 90% |
| DeepWalk | 59.04 | **68.25** | 69.75 | 42.82 | 45.49 | 45.57 | 54.22 | 61.91 | 0.62.11 | 81.60 | 86.13 | 86.82 |
| node2vec | 58.73 | 66.98 | 70.12 | 43.20 | 44.87 | 44.43 | 52.66 | 60.22 | 60.87 | 81.07 | 84.81 | 84.47 |
| Walklets | 58.17 | 65.61 | 66.68 | 42.16 | 46.83 | 49.09 | 52.57 | 59.25 | 60.96 | 79.45 | 84.20 | 84.59 |
| TADW | 19.25 | 32.69 | 46.27 | 48.10 | 49.25 | 48.98 | 25.52 | 56.51 | 67.92 | 54.28 | 54.43 | 54.35 |
| **t-PINE*** | **61.64** | 66.16 | **74.00** | **73.53** | **82.95** | **85.73** | **66.00** | **70.00** | **75.00** | **82.59** | **90.92** | **91.88** |
| Gain over DeepWalk | 4.4 | – | 6.1 | 71.7 | 82.4 | 88.1 | 21.7 | 13.1 | 20.8 | 1.2 | 5.6 | 5.8 |
| Gain over node2vec | 4.9 | – | 5.5 | 70.2 | 84.9 | 92.9 | 25.3 | 16.2 | 23.2 | 1.9 | 7.2 | 8.8 |
| Gain over Walklets | 6.0 | 0.8 | 11.0 | 74.4 | 77.1 | 74.6 | 25.5 | 18.1 | 23.0 | 4.0 | 8.0 | 8.6 |
| Gain over TADW | 220.2 | 102.4 | 59.9 | 52.9 | 68.4 | 75.0 | 158.6 | 23.9 | 10.4 | 52.2 | 67.0 | 69.0 |

### Q2. $K$-NN matrix Incorporation

**Setup.** As [5] argues, most real-world networks are sparse. Thus, the network structure is not fully reflected in the network adjacency matrix. In other words, if we merely employ the observed connections for representation learning, we can barely reconstruct the original network using the learned latent features. Therefore, a serious need to learn richer and more useful representations has emerged. In this paper, in addition to the adjacency matrix, $\mathbf{Y} \in \mathbb{R}^{V \times V}$, we utilize a $K$-NN matrix, $\mathbf{Z} \in \mathbb{R}^{V \times V}$, to leverage the quality of representations learned.

In accordance to that, we visualize the reconstructed graph of the **Terrorist** dataset, under the guidance of label information, as shown in Figure 2, using two different settings. First; (shown in Figure 2a) we form the tensor $\underline{\mathbf{X}}$ using only one view: the adjacency matrix. We decompose the tensor $\underline{\mathbf{X}}$ using CP decomposition. We then reconstruct the adjacency matrix using the two factor representations matrices, $\mathbf{A}$ and $\mathbf{B}$. Second; (shown in Figure 2b), we form the tensor $\underline{\mathbf{X}}$ using the two views: the adjacency and the $K$-NN matrices. We decompose the tensor $\underline{\mathbf{X}}$ using CP decomposition. We then reconstruct the graph, which includes the adjacency and the $K$-NN matrices-related information, using the two factor representation matrices, $\mathbf{A}$ and $\mathbf{B}$. The Terrorist dataset includes four classes of relationships that can connect any two terrorists together: colleagues, family, congregate, and contact.

**Results.** From Figure 2a, We observe that although the multi-label classification problem on the Terrorist dataset is a bit easier than on the other datasets, (it mostly exhibits homophily connectivity patterns), it is still insufficient to merely rely on immediate connections to clearly identify the four different communities. On the other hand, when we jointly exploit the adjacency and the $K$-NN views representations to reconstruct the graph, we are able to identify more pure and isolated communities with a less overlap, which is shown in Figure 2b.

### Q3-a. CP Decomposition and Predictability

**Setup.** Similar to interpretability, in the context of representation learning, obtaining highly predictable representations that effectively generalize across different downstream processes, without overfitting to the training data, has been to address [20]. Therefore, here in this experiment, we aim to thoroughly examine our claim that *the classifier's performance greatly depends on how well the nodes are*

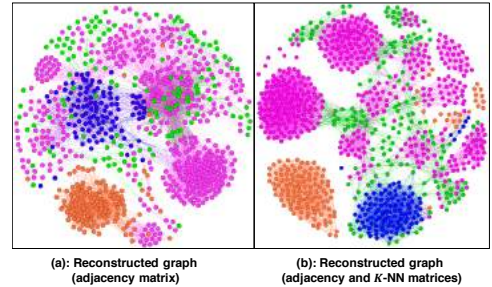

**(a): Reconstructed graph (adjacency matrix)**   **(b): Reconstructed graph (adjacency and $K$-NN matrices)**

**Figure. 2:** Visualization of the reconstructed graph of the **Terrorist** dataset, colored using label information. **(a):** Visualizes the reconstructed graph using the representations learned using adjacency matrix. **(b):** Visualizes the reconstructed graph using the representations learned using adjacency and $K$-NN matrices. The four communities are well-isolated and identified using the adjacency and $K$-NN matrices, which emphasizes the serious need to incorporate the $K$-NN matrix.

*separated in the feature space*, which would ultimately boost a simple classifier's (e.g., logistic regression), prediction accuracy. We use CiteSeer dataset to visualize the original vs. the reconstructed adjacency and $K$-NN views, as shown in Figure 3. The term "reconstructed" refers to the adjacency and $K$-NN views we approximate using decomposition factor matrices: $\mathbf{A} \in \mathbb{R}^{V \times d}$, $\mathbf{B} \in \mathbb{R}^{V \times d}$, and $\mathbf{C} \in \mathbb{R}^{L \times d}$. We reconstruct the adjacency view using $\mathbf{A}, \mathbf{B}$ and $\mathbf{C}(1,:)$, while we approximate the $K$-NN view using $\mathbf{A}, \mathbf{B}$ and $\mathbf{C}(2,:)$.

**Results.** Figures 3a and 3b visualize the adjacency view, $\mathbf{Y} \in \mathbb{R}^{V \times V}$, and the $K$-NN view, $\mathbf{Z} \in \mathbb{R}^{V \times V}$, which is processed using the $K$-NN algorithm using $K = 15$. Figures 3c and 3d visualize the reconstructed adjacency and $K$-NN views ($K = 15$). We infer that in contrast to the three classes, we can characterize using the original views (3a and 3b), CP decomposition successfully uncovers the six classes of CiteSeer dataset. As a result, we anticipate that using CP decomposition would significantly boost the classifier's predictability using feature representations in the low-dimensional space.

### Q3-b. CP Decomposition and Interpretability

**Setup.** Interpretability has been a must in almost all data mining and machine learning real-world applications [20]. In the context of representation learning, having easily interpretable representations is a very important, yet very challenging issue to address [20]. To this end, here in this paper, we compre-hensively examine t-PINE interpretable representations. Our interpretability experiment is driven by two motivations : (1) Demonstrate the "per-view" contribution to the representation learning process; and (2) Provide a proper guidance to eliminate meaningless, and useless dimensions. Utilizing the factor matrix $\mathbf{C} \in \mathbb{R}^{L \times d}$ generated by CP decomposition, we successfully achieves our experiment motivations.

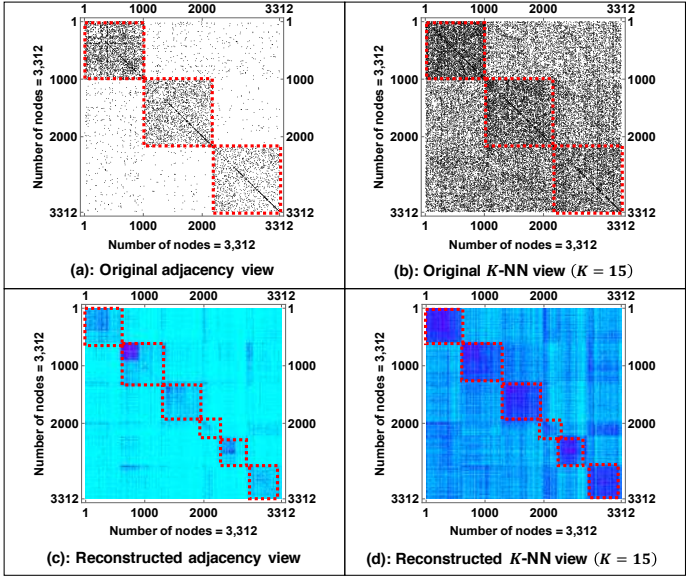

**Figure. 3:** Visualization of the original vs. the reconstructed adjacency and $K$-NN views on CiteSeer dataset, which has six classes. **(a):** Represents the visualization of the original adjacency view. **(b):** Shows the visualization of the original $K$-NN view ($K = 15$). **(c):** Represents the visualization of the reconstructed adjacency view. **(d):** Shows the visualization of the reconstructed $K$-NN view ($K = 15$). Reconstructed views using CP decomposition factor matrices helps to clearly identify the six classes compared to the three classes identified using original views of CiteSeer dataset.

*Motivation 1.* To further explore the role of the factor matrix $\mathbf{C} \in \mathbb{R}^{L \times d}$, we use the CiteSeer dataset. In our study, we have two views; the adjacency matrix, $\mathbf{Y} \in \mathbb{R}^{V \times V}$, and $K$-NN matrix, $\mathbf{Z} \in \mathbb{R}^{V \times V}$, which is processed using the $K$-NN algorithm using $K = 15$, and we set $d$ to a smaller value, 64, for ease of visualization. Similar conclusions should hold for other values of $d$. We choose the $K$ value that accompanied with the best performance on the CiteSeer dataset.

*Motivation 2.* In order to exclude unrelated dimensions, we arbitrarily set a weight-based threshold to $0.12$. Therefore, we remove any dimension with a weight less than the threshold. We use the same dataset and experimental settings we use in motivation 1.

**Results.**
*Motivation 1.* Figure 4 visualizes the weights of the of the 64 dimensions of the factor matrix $\mathbf{C} \in \mathbb{R}^{L \times d}$. We clearly can see that the adjacency view weights span over the bottom part of the figure, while the $K$-NN view weights spread over the top part of the figure. This leads us to conclude that the representation learning process not only impacted by the adjacency view, but it also influenced by the $K$-NN view, which is proven by the high weights assigned to this specific view.

*Motivation 2.* As shown in Figure 4, we remove 12 dimensions in total, and thus, we are left with 52 dimensions (the Micro-F1 score $= 64.31$) instead of 64 dimensions (the Micro-F1 score $= 65.51$). We don't observe a significant drop in accuracy after discarding the 12 useless dimensions. We also perform further analysis on the removed dimensions, and note that these low-weight dimensions are highly correlated with the remaining dimensions, which justifies the insignificant drop in accuracy.

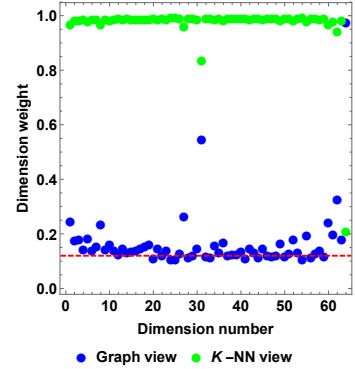

**Figure. 4:** Visualization of factor matrix $\mathbf{C} \in \mathbb{R}^{L \times d}$ weights on CiteSeer dataset. The bottom part represents the weights of the adjacency view of each representation dimension. The top part indicates the the weights of $K$-NN view of each representation dimension. The red-dashed line represents the threshold we set to .12 to discard low-weight, and useless dimensions. Any dimension with weight less than the threshold is discarded. Along with adjacency view, the $K$-NN view, drastically impacts the representation learning process.

### Q4. Parameter Sensitivity

**Setup.** The t-PINE requires two parameters: $K$, and $d$. Here we examine how t-PINE is sensitive, while varying $K$ from 2 to 40, and $\log_2 d$ ranges $3-9$, using the four evaluation datasets (shown in Table III). We perform 10-fold cross validation and report the mean Micro-F1 score results, when the training data is sparse ($10\%$).

**Parameter $K$.** In Figure 5a, we note that for the CiteSeer, Wikipedia, and Terrorist datasets, up to a certain level, the performance of t-PINE improves when $K$ increases, while it starts to either decrease or maintain the same performance afterward. On the other hand, the WebKB dataset has a more stable performance with respect to $K$. In the interest of space, we omit the results of Macro-F1 score, because they follow the exact same trend.

**Parameter $d$.** For a fair comparison, in Figure 5b, we perform sensitivity analysis for the whole baseline methods using WebKB dataset. We observe that by only using a $10\%$ of labeled data, t-PINE extraordinarily outperforms the state-of-the-art techniques for different dimensions. Further, we perceive that up to $d = 128$ or $\log_2 d = 7$, t-PINE's performance consistently improves, and then it starts to drop. This can be attributed to the fact that including extra dimensions can lead to introducing some noise to the representation feature space, which ultimately would negatively affect the performance. A similar trend, though less performance, is observed for the baselines as well. Due to space constraints, we omit the results on other datasets, where a similar conclusion holds, and the Marco-F1 score results.

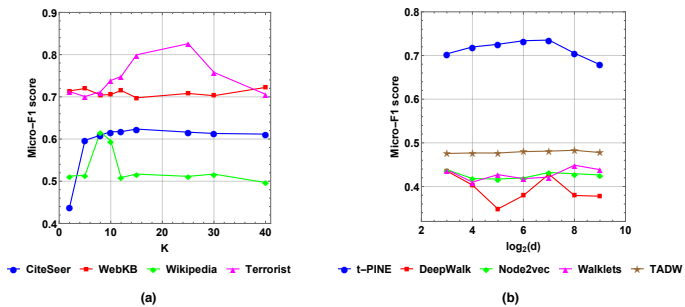

**Figure. 5:** Sensitivity analysis of t-PINE method using four real-world datasets. **(a):** Sensitivity analysis with respect to $K$, where $K$ ranges $2-40$. Stable performance in the range $15-40$, for most datasets. The only exception is Terrorist dataset that has increased performance in that range. **(b):** Sensitivity analysis with respect to $\log_2 d$ using WebKb dataset. t-PINE significantly outperforms the state-of-the-art methods even when labeled data is sparse.

## VI. Conclusion

We propose a novel, effective, and interpretable representation leaning method, t-PINE. It employs multi-view graph information by jointly exploiting the conventional adjacency view along with its corresponding side information view: $K$-NN matrix. t-PINE comprises two steps: (1) systematic exploration of higher-order proximity step using the intuition of $K$-NN algorithm, and (2) profound representation learning step using CP decomposition. We extensively evaluate our proposed method using four real-world datasets on different multi-label classification problems with respect to different aspects, i.e., *efficacy*, *interpretability*, and *predictability*. We contrast t-PINE's performance against four representation learning state-of-the-art methods. Empirical demonstrations show that t-PINE outperforms baseline techniques by up to $158.6\%$ with respect to Micro-F1, when the labeled data is sparse. In addition, we observe that using a multi-view information graph enhances the identification of multiple classes, while leveraging CP decomposition dramatically improves the interpretability and prediction capability of representation learning. Interestingly, we note that t-PINE constantly outperforms the most competitive baseline method, TADW, which leverages side information. In our future work, we will address the issue of embedding update, especially for a recently-joined node that has no evident connections.